\documentclass[letterpaper, 10 pt, conference]{ieeeconf}  

\IEEEoverridecommandlockouts
\usepackage{caption}
\usepackage{graphics}
\usepackage{epsfig} 
\usepackage{mathptmx}
\usepackage{times}
\usepackage{amsmath}
\usepackage{amssymb} 
\usepackage{algorithm}
\usepackage{multirow}
\usepackage{tabularx}
\usepackage{colortbl}
\usepackage[table,xcdraw]{xcolor}
\usepackage{booktabs}
\usepackage{amsmath}
\usepackage{makecell}
\usepackage{algpseudocode}
\usepackage{tcolorbox}
\usepackage{comment}
\let\labelindent\originallabelindent
\usepackage[leftmargin=1.0em]{quoting}
\usepackage{enumitem}

\title{\LARGE \bf
Adapting Robot's Explanation for Failures Based on Observed Human Behavior in Human-Robot Collaboration
}
\author{Andreas Naoum$^{1*}$, Parag Khanna$^{1*}$, Elmira Yadollahi$^{2}$, Mårten Björkman$^{1}$, and Christian Smith$^{1}$
\thanks{$^*$ These authors contributed equally. 
$^{1}$Division of Robotics, Perception and Learning (RPL), EECS, KTH Royal Institute of Technology, Sweden \tt\{anaoum, paragk, celle, ccs\}@kth.se}%
\thanks{$^{2}$Lancaster University
        {\tt e.yadollahi@lancaster.ac.uk}}%
}
\begin{document}

\maketitle
\thispagestyle{empty}
\pagestyle{empty}

\begin{abstract}
This work aims to interpret human behavior to anticipate potential user confusion when a robot provides explanations for failure, allowing the robot to adapt its explanations for more natural and efficient collaboration. Using a dataset \cite{khanna_dataset_explanations} that included facial emotion detection, eye gaze estimation, and gestures from 55 participants in a user study \cite{exp_strategies_roman_khanna2023}, we analyzed how human behavior changed in response to different types of failures and varying explanation levels. Our goal is to assess whether human collaborators are ready to accept less detailed explanations without inducing confusion.
We formulate a data-driven predictor to predict human confusion during robot failure explanations. We also propose and evaluate a mechanism, based on the predictor, to adapt the explanation level according to observed human behavior. The promising results from this evaluation indicate the potential of this research in adapting a robot's explanations for failures to enhance the collaborative experience. 
\end{abstract}
\section{Introduction}
Recent advancements in robotics have enabled robots to collaborate with humans in a variety of tasks \cite{cobots_advances}. However, the uncertain nature of environments in which robots operate often leads to failures \cite{failures_hri,robot_safety}. In instances where robots encounter failures, human intervention in certain cases can easily troubleshoot the problem efficiently and effectively \cite{exp_strategies_roman_khanna2023,failures_hri,better_faulty}. Therefore, a crucial aspect of this collaboration is the robot's ability to communicate when a failure occurs, explain why the failure happened, and, if possible, suggest a course of action for resolution. This communication ability is not only essential for successful collaboration but also for building rapport and trust \cite{Correia_fault_justification, What_If_It_Is_Wrong_Karli2023} in robots.

Several studies have explored the content and automated generation of failure explanations to improve communication in human robot collaboration (HRC) \cite{xai_failures, why_fail}. Effective collaboration requires robots to optimize efficiency and adapt to the needs of their human counterparts \cite{adaptivity_hri}. A pivotal factor in this dynamic is the ability of robots to identify and manage user confusion during interactions. Confusion arises when users encounter information or tasks that challenge their current understanding or skills \cite{detecting_confusion_int_2021}. When confusion is properly managed or resolved, it can motivate critical thinking and cognitive engagement. On the other hand, if confusion is left unmanaged, it can lead to frustration, errors, and disengagement, ultimately hindering both the interaction experience and user satisfaction. The fine line between these two outcomes highlights the complexity of identifying and managing confusion effectively in HRC. 

In addressing confusion during recurring failure scenarios, it is essential for robots to tailor their failure explanations based on the user's mental state and familiarity with the issue. There are instances when less information is sufficient to convey a failure, particularly if the collaborator is already aware of the problem. While succinct failure explanations can streamline communication, they may inadvertently induce confusion and trigger negative emotional responses \cite{detecting_confusion_int_2021, identify_int_confusion_LI2023, taxonomy_social_errorsin_HRI_jrHRI2021}. Emotional cues, combined with non-verbal signals like gestures and eye gaze, can reveal that the human collaborator is not adequately prepared for a decrease in explanatory detail during failure situations. Therefore, the primary aim of this work is to explore how robots can effectively observe human behavior to adapt failure explanations, thereby minimizing confusion and enhancing collaborative outcomes in HRC scenarios of recurring failures.

\subsection{Contributions}
This paper builds upon our previous work \cite{exp_strategies_roman_khanna2023}, where we explored explanation strategies for recurring failure scenarios in HRC. Using the multimodal data from that study \cite{zenodo_dataset_reflex}, we are able to identify and predict confusion induction in failure explanation instances, allowing for proactive mitigation of negative reactions. We posit that the integration of an adaptive mechanism for delivering failure explanations will improve the efficiency and fluidity of HRC. 
\begin{figure}[t]
      \centering
      \setlength\abovecaptionskip{-0.7\baselineskip}
      \includegraphics[scale=0.4,trim={0.5cm 3cm 0cm 0.1cm},clip]{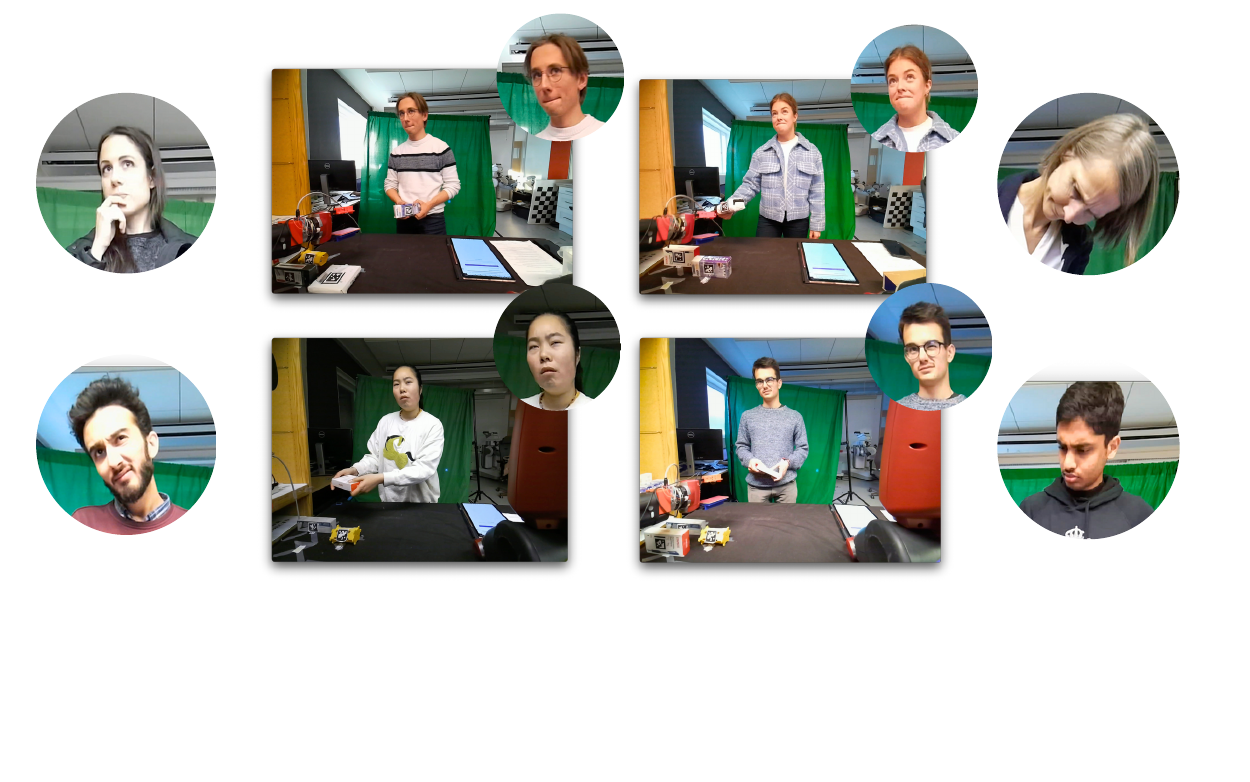}
      \caption{Human Confusion: This figure illustrates cases of confusion during robotic explanations for a robotic failure.}
      \label{figure1}
      \vspace{-3.0mm}
\end{figure}
The main contributions of this work are:
\begin{itemize}[leftmargin=*]
    \item We introduce an automated approach to assess confusion induction in failure instances.
    \item We develop a data-driven predictor to predict confusion induction in humans when a robot collaborator provides explanations for failures, based on the human behavioral response to the robotic failures.
    \item We propose and evaluate a mechanism to adapt the level of detail of explanations for robot failures based on observed human behavior, utilizing the aforementioned predictor.
\end{itemize}
\section{Background} 
\subsection{Confusion in Human-Machine Interaction}
At its most basic level, confusion can be understood as ``lack of clarity.'' Although confusion has been extensively studied in social sciences and psychology, research on confusion in HRI remains limited. According to Na Li et al \cite{detecting_confusion_int_2021}, confusion in human-machine interaction can be defined as:
\begin{quoting}
``Confusion is a mental state where under certain circumstances, a human experiences obstacles in the flow of interaction. A series of behavior responses (which may be nonverbal, verbal, and, or non-linguistic vocal expression) may be triggered, and the human who is confused will typically want to solve the state of cognitive disequilibrium in a reasonable duration. However, if the confusion state is maintained over a longer duration, the interlocutor may become frustrated, or even drop out of the ongoing interaction.''
\end{quoting}
When specific scenarios elicit a state of confusion, this phenomenon is referred to as ``Confusion Induction''.
Literature \cite{detecting_confusion_int_2021,identify_int_confusion_LI2023} identifies four types of confusion induction: complex information, contradictory information, insufficient information, and feedback inconsistencies. In our case, insufficient information is especially critical. This occurs when essential details are missing, preventing effective understanding and decision-making and leading to uncertainty. 

\subsection{Facial Emotional Analysis}
Facial emotional analysis is a method relying on facial expressions to interpret human emotions \cite{emotion_hri}. Early research \cite{basic_emotions} identified six universal emotions consistently recognized through these expressions across various cultures. These expressions also convey broader emotional dimensions like arousal and valence. Recent research \cite{emotions28, HUME_2024_deep_learning_Facial_Exp, emotions_intersectionality} challenges the traditional views, suggesting that facial expressions are complex and high-dimensional, and cannot be easily reduced to a few basic categories. 
A study on naturalistic facial expressions revealed that perception of these expressions involves at least 28 distinct dimensions of meaning \cite{emotions28}. Further, a deep neural network \cite{HUME_2024_deep_learning_Facial_Exp} predicting the culturally specific meanings people assign to facial movements, while accounting for physical appearance and context, identified 28 distinct dimensions of facial expression. Given this complexity, focusing on high-dimensional emotions, offers a more complete and nuanced interpretation of human behavior than relying solely on basic emotions \cite{emotions28, barrett2019emotional}.

\section{Related Work}
This research is part of a broader effort to make HRI more intuitive by observing human behavior. Significant progress has been made with models that analyze various data types, such as voice tone, facial expressions, and body language, to detect robot errors \cite{Analysis_of_human_rxn_to_robot_convo_failures_DIMOS2021, Model_human_response_to_robot_Error_Stiber2022}, identify human states \cite{identify_int_confusion_LI2023, Human_Affective_State_Recognition_jirak2022} and predict human reactions \cite{Multimodal_User_Feedback_AGNES2022, goswami2020socialengagingpeer}. However, much of this work has emphasized external indicators without fully integrating the emotional context underlying these behaviors. 

In \cite{cooadaptive_HRI_Sofie2022}, the authors emphasized the crucial role of robots' social adaptation abilities in building trust through their capacity to adapt to human needs and emotions. Although recent studies have explored human states like engagement \cite{UE_HRI_dataset_user_engangeinHRI_2017}, or disengagement \cite{goswami2020socialengagingpeer}, the investigation of confusion in HRI has received less attention. Na Li et al. studied confusion in user-avatar dialogues \cite{detecting_confusion_int_2021} and in HRI \cite{identify_int_confusion_LI2023} through facial emotion analysis, eye gaze, and head pose, identifying relationships between these indicators and states of confusion. Despite these advancements, their approach is limited by traditional views of emotions.  

In parallel, multiple studies in the areas of HRI and collaboration have focused on how failures can affect trust and perception \cite{Correia_fault_justification,van2017lack}, and ways to minimize these negative effects by providing explanations for robot behavior \cite{das2021explainable,exp_strategies_roman_khanna2023}.
For explainability in robot failures, a study \cite{das2021explainable} investigated the types of explanation that helped non-experts to identify robot failures and assist the recovery via introducing failure explanation, and found explanations including the context and history of previous actions to be most effective.
It was shown in \cite{alvanpour2020robot} 
that an explanation of predicted faults could contribute to the efficiency of designing the robot and avoiding future failures.
Effects of different types and amounts of explanation were discussed in \cite{XAI-Linder_effects_Exp_levels}, where detailed explanations improved human understanding and performance but required more time and attention to process. In our previous work \cite{exp_strategies_roman_khanna2023}, various explanation levels and explanation progression strategies to handle recurring robotic failure scenarios in HRC were explored. 
The findings suggested that for complex failures, we can reduce the information provided in the explanation by initially using a high explanation level, thereby minimizing repetitions and collaboration times.

We address the limitations of prior work by incorporating better interpretations of human behavior through rich emotional analysis and specific physiological signals associated with the emotional state into the prediction models for human reactions. Further, by predicting emotional reactions, robots could better decide when to adjust explanation levels, adapting responses to the user's emotional and cognitive needs, rather than relying on predetermined strategies.
\section{Data Collection and Analysis}
\vspace{-1.0075mm}
\subsection{Data Sources}
The data \cite{zenodo_dataset_reflex} utilized for this work was gathered through a user study \cite{exp_strategies_roman_khanna2023, khanna2023userstudyexploringrole}. The study involved 55 participants who were given the task of placing household objects on a table while a Baxter robot was responsible for picking them up and placing them on a shelf. For each object, the robot executed the following actions: detect the object, 'Pick' it up, 'Carry' it, and finally
'Place' the object at a desired location in the shelf. 
The study happened in four rounds with four objects in each round.
Throughout the study, the participants interacted with the robot that encountered pre-programmed 11 failures while performing the 'Pick,' 'Carry,' and 'Place' actions.
The robot needed human help in resolving these failures.
When the robot failed to pick an object since it did not fit in its gripper, the human handed over the object to the robot by placing it in the gripper.
When the robot failed to carry an object due to its load or failed to place an object as the desired location is beyond its reach, the robot handed over the object to the human, who would carry and place the object on the shelf.
The reader is advised to see the supplementary video for a visual representation of the robotic failures and resolution by humans; and the related works \cite{exp_strategies_roman_khanna2023,khanna2023userstudyexploringrole} for further details about the study. 
For data collection, in the absence of a relevant ethics board, we followed guidelines of the Declaration of Helsinki. Participants began
by completing a consent form for data collection. Particularly, they consented to the use and distribution of their anonymized data and the use of collected visual data in academic articles and presentations. 
The robot provided failure explanations and resolution actions to the participants in order to collaboratively resolve the failure and successfully complete the task.
\begin{table}[b]
    \centering
    \setlength\abovecaptionskip{-0.15\baselineskip}
    \caption{Explanation Strategies from \cite{khanna2023userstudyexploringrole}}
    \label{tab:round-results}
    \resizebox{0.7\columnwidth}{!}{
    \begin{tabular}{clcccc}
        \hline
        ID & Details & Round 1 & Round 2 & Round 3 & Round 4 \\
        \hline
        C1 & Fixed-Low & Low & Low & Low & Low \\
        C2 & Fixed-Medium & Mid & Mid & Mid & Mid \\
        C3 & Fixed-High & High & High & High & High \\
        D1 & Decay-Slow & High & Mid & Low & None \\
        D2 & Decay-Rapid & High & Low & Low & Low \\
        \hline
    \end{tabular}
    }
\end{table}
The robot employed four distinct levels of explanation to assist participants \cite{exp_strategies_roman_khanna2023}: 
\begin{itemize}[leftmargin=*]
    \item \textbf{Level Zero (Non-verbal):} The robot used only non-verbal cues, such as head shakes and the failure action itself, to indicate failures.
    \item \textbf{Level Low (Action-based from \cite{chakraborti2020emerging}):} The robot verbally described the failure and suggested a resolution based on the specific action that caused the issue. (
    ``I can’t pick up the object'', ``Hand it to me.'')
    \item \textbf{Level Medium (Context-based from \cite{chakraborti2020emerging}):} The robot provided an explanation of the failure, including its cause, followed by a recommended resolution. (``I can’t pick up the object because it doesn’t fit in my gripper.'', ``Can you hand it over to me?'')
    \item \textbf{Level High (Context+History-based from \cite{chakraborti2020emerging}):} The robot offered a detailed explanation that included references to past successful actions and the cause of failure, along with a step-by-step guide for resolving the failure. (``I can detect the object, but I can’t pick it up because it doesn’t fit in my gripper.'' ``Can you hand it over to me by placing it in my gripper?'')
\end{itemize}
Table \ref{tab:round-results} outlines the five experimental conditions applied across different rounds of the study, with each condition representing an explanation variation strategy involving varying levels of explanation, 11 participants each for a condition. There are 2 categories: ``Fixed", keeping the explanation constant across each round, and ``Decay", decreasing the explanation starting from a high level.
\subsection{Human Reaction Modeling}
\begin{figure}[t]
      \centering
      \setlength\abovecaptionskip{-0.05\baselineskip}
      \includegraphics[scale=0.22,trim={6.0cm 15.55cm 5.0cm 0.75cm},clip]{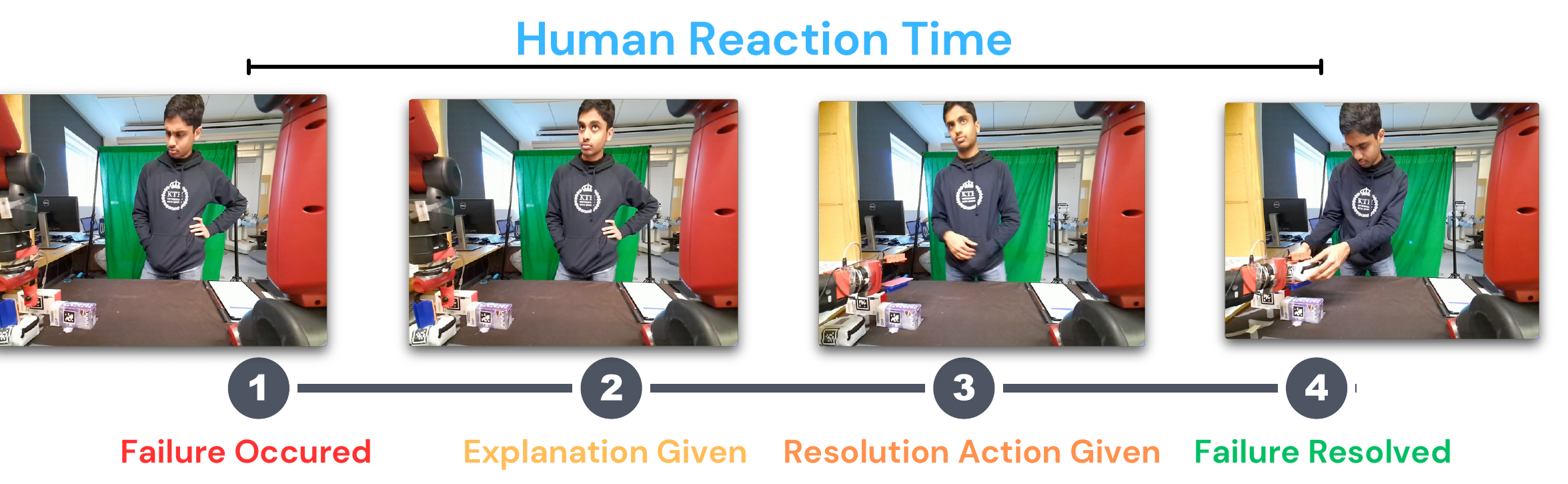}
    \caption{Human Reaction Modelling for Failure Explanation}
      \label{figurelabel}
      \vspace{-4.0mm}
\end{figure}
Inspired by the recent study in \cite{What_If_It_Is_Wrong_Karli2023, Model_human_response_to_robot_Error_Stiber2022}, we divided the participants’ behaviors into four distinct phases, each representing a key interaction point for every failure event. 

\begin{itemize}[leftmargin=*]
    \item \textbf{Pre}: The phase before the failure occurs.
    \item \textbf{Failure}: The phase when the failure happens.
    \item \textbf{Explanation}: The phase where the robot communicates the reasons behind the failure.
    \item \textbf{Resolution}: The phase where the robot asks the participant to take steps to resolve the failure.
\end{itemize}

\subsection{Data Extraction and Processing}
For the analysis of participants' behavior \cite{khanna_dataset_explanations}, we consider:

\subsubsection{\textbf{Facial Emotions}}
We utilized Hume's \cite{HUME_2024_deep_learning_Facial_Exp,HUME_API} expression measurement API to identify facial emotions, which included 48 emotions \cite{khanna_dataset_explanations}. Each emotion was scored on a scale from 0 to 1, representing the likelihood ($\mathcal{L}$) that a human would perceive that emotion in the facial expressions. We computed both average and peak emotional scores during each phase and the changes between phases. To identify the most relevant emotions, we analyzed their frequency of occurrence and variation in $\mathcal{L}$ across phases. We narrowed down the emotions with high probability of occurrence and notable changes, listed in Table \ref{tab:behavior_measures}.


\subsubsection{\textbf{Eye Gaze}}
We employed OpenFace \cite{openface} for eye-tracking technology to measure the direction and duration of participants' gaze. We categorized the gaze into three main focal points: the task, the robot, and miscellaneous. The percentage of time spent looking at each focal point was calculated for attention distribution. 

\subsubsection{\textbf{Gestures}}
Hand-to-head/face gestures were detected using the Hume API \cite{HUME_API}, while head tilting was detected using OpenFace head pose estimation \cite{openface}.
\begin{table}[b]
\setlength\abovecaptionskip{-0.25\baselineskip}
\caption{Behavior Measures}
    \centering
    \resizebox{0.6\columnwidth}{!}{
    \begin{tabular}{p{3cm} p{3cm}}
        \toprule
        \textbf{Category} & \textbf{Indicators}  \\
        \midrule
        \multirow{4}{4cm}{Negative Emotions \\ (Avg/Max)}
        & Confusion, Doubt \\
        & Disappointment, Anxiety  \\
        & Anger, Distress \\
        & Surprise (Negative) \\
        \midrule
        \multirow{2}{4cm}{Positive Emotions \\ (Avg/Max)}
        & Satisfaction, Interest \\
        & Contentment, Desire \\
        \midrule
        \multirow{3}{4cm}{Eye Gaze \\ (\%)}
        & Gaze on Robot \\
        & Gaze on Task \\
        & Gaze on Misc. \\
        \midrule
        \multirow{2}{4cm}{Gestures \\ (Presence)}
        & Hands on Head/Face  \\
        & Head Tilting \\
        \bottomrule
    \end{tabular}
    }
    \label{tab:behavior_measures}
     \vspace{-1.mm}
\end{table}

\begin{figure*}[thpb]
    \centering
    \setlength\abovecaptionskip{-0.04\baselineskip}
    \begin{minipage}{\textwidth}
        \centering
        \setlength\abovecaptionskip{-0.05\baselineskip} \includegraphics[width=1.0\textwidth,height=0.3\textheight,keepaspectratio,trim={0.0cm 0.6cm 0.0cm 0.1cm},clip]{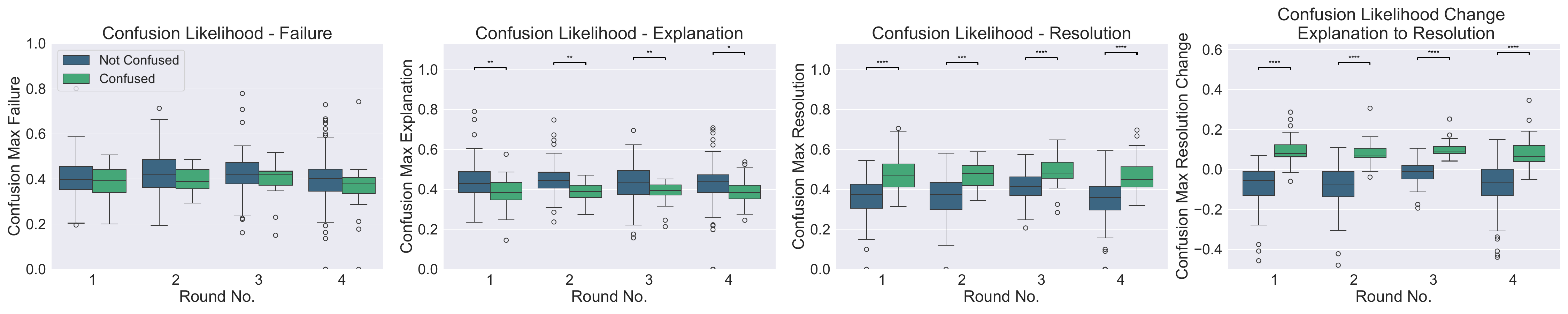}
        \caption*{(a)}
    \end{minipage}
\vspace{-1.5em}
    \vskip\baselineskip
    \hspace{-4.0em}
    \begin{minipage}{0.24\textwidth}
        \includegraphics[width=\linewidth,height=0.13\textheight,trim={0.85cm 0.0cm 0.5cm 0.1cm},clip]{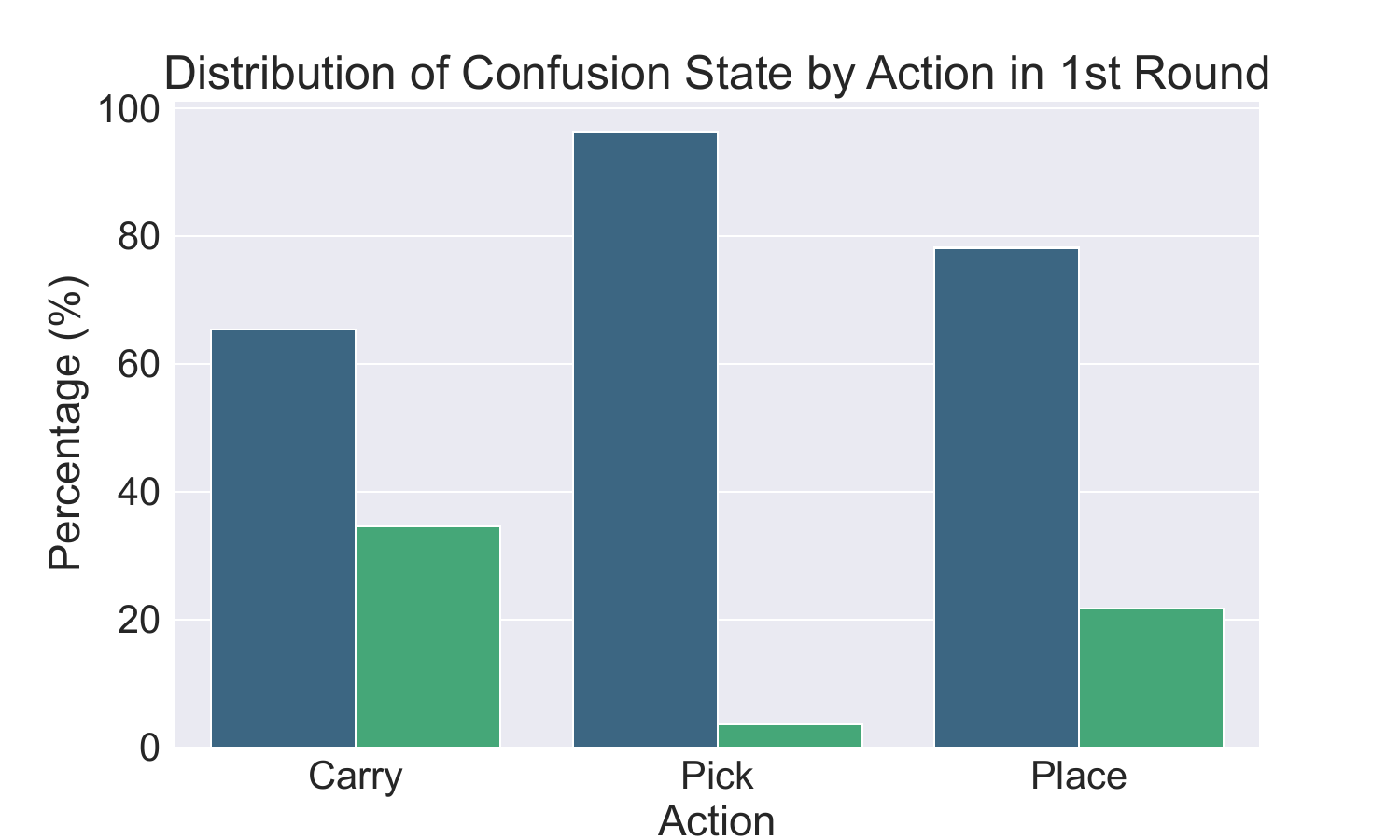}
        \caption*{(b)}
    \end{minipage}%
    \hspace{-0.8em}
    \begin{minipage}{0.24\textwidth}
        \includegraphics[width=\linewidth,height=0.13\textheight,trim={0.85cm 0.0cm 0.5cm 0.1cm},clip]{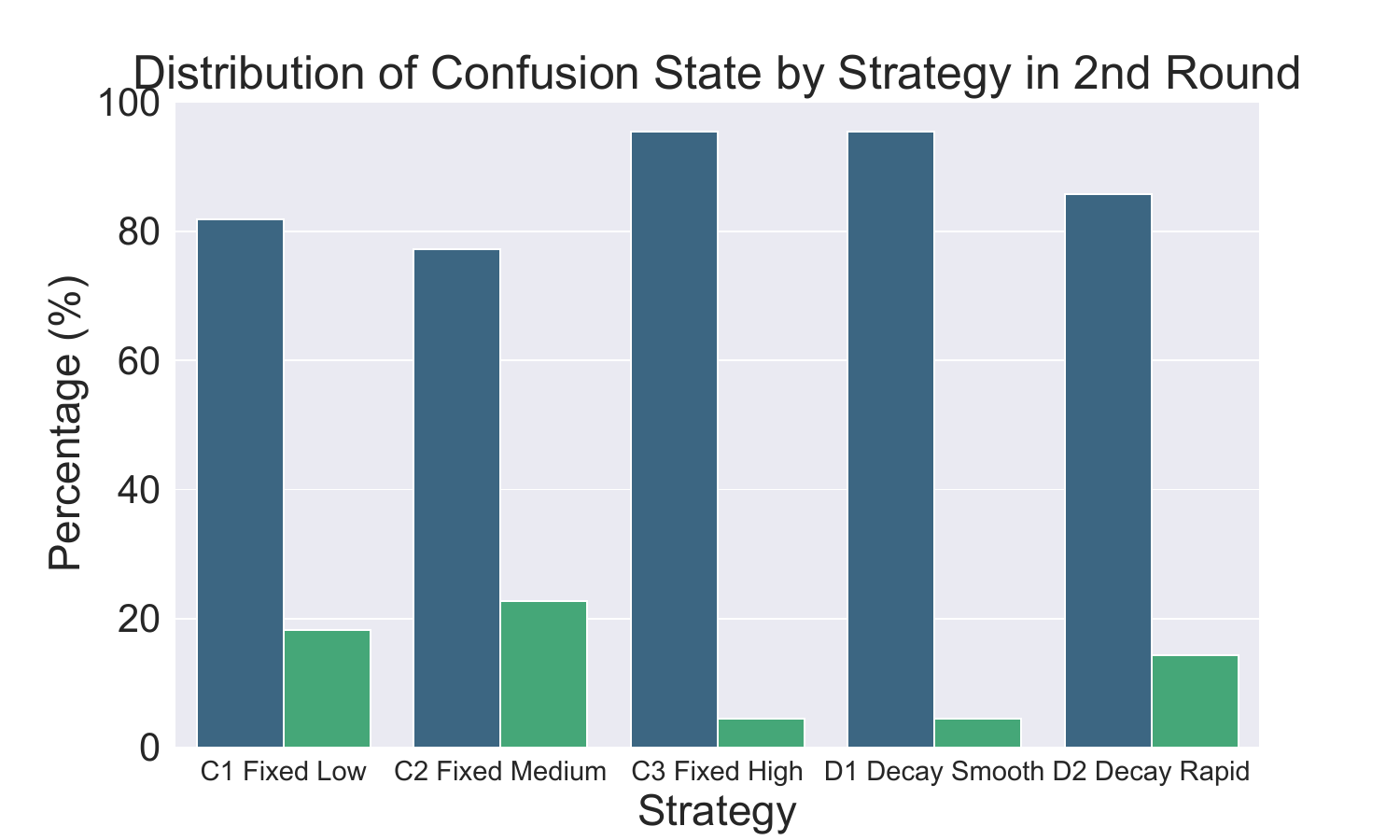}
        \caption*{(c)}
    \end{minipage}%
    \hspace{-0.6em}
    \begin{minipage}{0.49\textwidth}
        \includegraphics[width=1.25\linewidth,height=0.13\textheight,trim={4.4cm 0.4cm 0.5cm 0.1cm},clip]{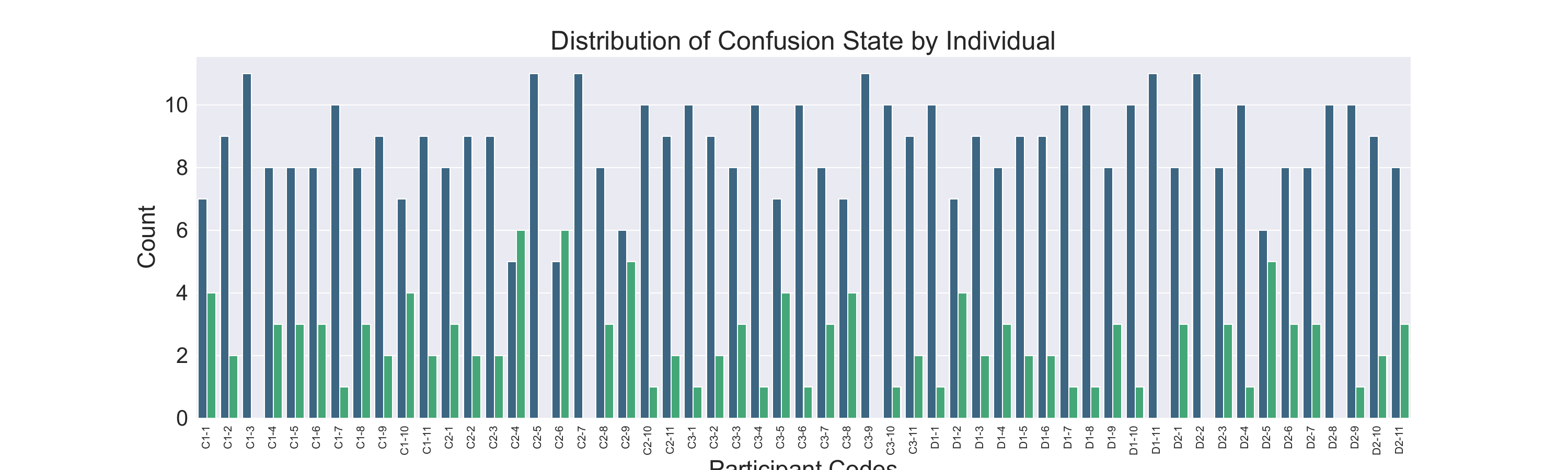}
        \caption*{(d)}
    \end{minipage}

    \caption{Results: (a) $\mathcal{L}_c$ values in different phases, Confusion States Variation by (b) Actions (c) Strategies (d) Individuals}
    \label{fig:Data_analysis}
\end{figure*}
\subsection{Identify Confusion Induction}
\begin{algorithm}[t]
\small	
\caption{Identifying Confusion Induction}
\begin{algorithmic}[1]
\Function{set\_confusion}{confusion}
    \If{\Call{high\_confusion}{confusion}}
        \State \Return \textbf{``Confused"}
    \ElsIf{\Call{persistent\_confusion}{confusion}}
        \State \Return \textbf{``Confused"}
    \Else
        \State \Return \textbf{``Not Confused"}
    \EndIf
\EndFunction

\Statex \textbf{Initialize} \textit{t\_high}, \textit{t\_change}.
\Function{high\_confusion}{confusion}
\State \hspace{1em} Return If $\mathcal{L}_c$ at resolution $>$ \textit{t\_high}, \textbf{True}; else, \textbf{False}.
\EndFunction

\Function{persistent\_confusion}{confusion}
\State \hspace{1em} a. If $\mathcal{L}_c$ increased by \textit{$t_{\text{change}}$} at explanation and $\mathcal{L}_c$ is not reduced by \textit{$t_{\text{change}}$} at resolution, return \textbf{True}.
\State \hspace{1em} b. If $\mathcal{L}_c$ increased by \textit{$t_{\text{change}}$} at failure and $\mathcal{L}_c$ is not reduced by \textit{$t_{\text{change}}$} at resolution, return \textbf{True}.
\State \hspace{1em} c. If $\mathcal{L}_c$ increased by \textit{$t_{\text{change}}$} at resolution, return \textbf{True}.
\State \hspace{1em} d. Return \textbf{False}.
\EndFunction
\end{algorithmic}
\label{alg:1}
\end{algorithm}

Prior research \cite{detecting_confusion_int_2021} suggests that participants may not always be aware of their own confusion, even in situations specifically designed to induce it. Interestingly in \cite{detecting_confusion_int_2021}, there was no significant relationship between self-reported confusion scores and induced confusion states, but a relationship was found between physical states and induced confusion. These findings highlight a potential limitation of relying on self-reporting and underscore the need for more objective measures \cite{automatic_assessment}. Therefore, we decided to use automated approach based on facial emotion analysis with a specific focus on confusion to capture subtle indicators objectively.

We developed our algorithm, Algorithm \ref{alg:1}, to identify confusion induction based on key interaction points through a systematic approach, combining insights from existing literature with empirical testing. The likelihood of confusion ($\mathcal{L}_c$) values used in the algorithm are derived from the facial emotion recognition mentioned earlier. Notably, confusion is recognized as one of the emotions that exhibits strong evidence of universality across different cultures \cite{HUME_2024_deep_learning_Facial_Exp}, a characteristic that enhances the algorithm’s robustness. By applying consistent criteria across all participants, our approach ensures a uniform and objective analysis, reducing the variability that can sometimes arise from self-reporting or manual annotation \cite{detecting_confusion_int_2021, automatic_assessment}. 

When participants encounter a high or persistent level of confusion, this results in what we term ``Confusion Induction." For simplicity, we will refer to this condition as being ``Confused". The algorithm is designed to identify such instances and operates as follows: First, it assesses whether the participant exhibits a high $\mathcal{L}_c$ during the resolution phase, referred to as ``High Confusion.'' If this is the case, the participant is classified as ``Confused.'' If not, the algorithm then checks for an unresolved increase in $\mathcal{L}_c$, termed ``Persistent Confusion.'' The rationale behind this is that the participant should not remain confused after the resolution action is provided. If there is an increase in $\mathcal{L}_c$ during the failure or explanation phases and an insufficient decrease after the resolution, the participant is also classified as ``Confused.'' Finally, if an increase in $\mathcal{L}_c$ is observed during the resolution phase itself, this too results in a classification of ``Confused.''

Algorithm \ref{alg:1} employs two key parameters: the high threshold ($t_{\text{high}}$), set at 0.7, and the change threshold ($t_{\text{change}}$), set at 0.05; and we find that the algorithm is not very sensitive to these values. 
The $t_{\text{high}}$ is used as the initial measure to identify high $\mathcal{L}_c$, while the $t_{\text{change}}$ plays a crucial role in identifying subtle yet important changes in the $\mathcal{L}_c$. Both were fine-tuned through direct observation and iterative testing. Since there is no universal threshold and different use cases may require varying sensitivity levels \cite{HUME_API,HUME_2024_deep_learning_Facial_Exp}, the thresholds were set based on our observations and are suitable for our case while remaining adaptable for similar contexts. Specifically, there were few instances that exceeded the $t_{\text{high}}$ in the resolution phase, and direct observation confirms that $t_{\text{high}}$ effectively captured those instances. Regarding the $t_{\text{change}}$, the algorithm uses the previous phase as a baseline for comparison of the $\mathcal{L}_c$ values to determine whether there was an increase or not. To determine if there is a sufficient decrease, the algorithm compares the $\mathcal{L}_c$ from the phase where the increase occurred with the $\mathcal{L}_c$ at the resolution. To effectively capture significant changes in $\mathcal{L}_c$, $t_{\text{change}}$ value is set above the level of random noise in the detected $\mathcal{L}_c$, ensuring only meaningful changes in $\mathcal{L}_c$ are detected, filtering out normal fluctuations. Empirical testing showed that this approach reliably distinguishes between insignificant variations and actual changes in $\mathcal{L}_c$. 

We use Algorithm \ref{alg:1} to identify confusion induction in our dataset. 
Fig. 3(a) demonstrates $\mathcal{L}_c$ across the failure, explanation, and resolution phases, and the change in $\mathcal{L}_c$ from explanation to resolution.
Comparing ``Confused" and ``Not confused" states over the four rounds reveals interesting observations supporting the algorithm.
Certain ``Not confused" instances maintain their $\mathcal{L}_c$, while others experience an increase in $\mathcal{L}_c$ during the failure or particularly the explanation phase, followed by a 
decrease in $\mathcal{L}_c$  during the resolution phase, indicating a pattern of productive confusion \cite{detecting_confusion_int_2021,identify_int_confusion_LI2023}.
In contrast, most ``Confused" instances show no increase in $\mathcal{L}_c$ during the failure and explanation phases, yet exhibit a notable rise in $\mathcal{L}_c$ during the resolution phase, resulting in a few very high values. Few ``Confused" instances show an increase during failure and explanation but never manage to decrease it. This distinct pattern suggests that confusion persists or intensifies for ``Confused" instances, even as resolution action was given, leading participants to be uncertain about the next steps. Also, there is a significance difference in $\mathcal{L}_c$ during the explanation phase, with ``Not confused" state displaying higher $\mathcal{L}_c$. The difference gets more pronounced but reversed during the resolution phase, with ``Confused"  state showing a high $\mathcal{L}_c$.

\subsection{Observations for Confusion Induction } 

The analysis of the extracted data suggests that confusion
is closely tied to the complexity of the action, the
explanation strategies employed, and the participant. Specifically:

\subsubsection{\textbf{Distribution of Confusion State by Action}}
Fig. \ref{fig:Data_analysis}(b) shows the distribution of confusion states based on the action in the initial round. The actions ``Place" and ``Carry" are associated with a high percentage (20-40\%) of ``Confused" instances, while ``Pick" action has the lowest. This indicates that participants perceived ``Place" and ``Carry" actions as more complex than the ``Pick" action. This interpretation aligns with findings from \cite{khanna2023userstudyexploringrole}, which reported a 100\% success rate for the “Pick” action across all levels in the initial round, whereas “Place” and “Carry” resulted in lower success rates, further indicating their relative difficulty.

\subsubsection{\textbf{Distribution of Confusion State by Strategy}}
Fig. \ref{fig:Data_analysis}(c) categorizes confusion states by different strategies in the second round. The ``Fixed High'' and ``Decay Smooth'' strategies resulted in the highest percentage of ``Not Confused” instances, indicating it was most effective in avoiding confusion induction. Conversely, the ``Fixed Low” and ``Fixed Medium" strategies had the highest percentage of ``Confused” instances, suggesting it may have been less effective in maintaining participant clarity. Additionally, the “Decay Rapid” strategy led to lower confusion compared to the “Fixed Low” and “Fixed Medium”. Notably, all strategies that initially provided a high level of explanation resulted in less confusion in the second round, further supporting the conclusion in \cite{khanna2023userstudyexploringrole} that users respond better to lower explanation levels after being exposed to higher levels earlier.

\subsubsection{\textbf{Distribution of Confusion State by Individual}}
  There is a clear variability in how individuals experience confusion, with some being consistently ``Not Confused” and others being more frequently ``Confused" (Fig. \ref{fig:Data_analysis}(d)). This highlights the individual differences in handling failure explanations and the potential need for adaptable strategies in HRC.

\section{Predicting Confusion Induction}
\begin{table}[t]
\centering
\setlength\abovecaptionskip{-0.15\baselineskip}
\caption{Features Used for Model Training}
\begin{tabular}{|m{2.5 cm}|m{5cm}|} 
\hline
\textbf{Feature Category} & \textbf{Description} \\ \hline
\textit{Action (A) } & The specific action performed by the robot. \( A = \{ \text{Pick}, \text{Place}, \text{Carry} \} \)\\ \hline
\textit{Explanation Decrease (D) } & Indicating if the explanation level decreases. D = 1 if true, else D = 0\\ \hline
\textit{Last Reaction (X) } & Data from the participant's last reaction to the same task in explanation and resolution phases (maximum, average and change in positive and negative emotions, eye gaze, and gestures)\\ \hline
\textit{Failure Reaction (X) } & Data from the participant's reaction during the current failure phase (maximum, average and change in positive and negative emotions, eye gaze, and gestures)\\ \hline
\end{tabular}
\label{tab:features}
\vspace{-4mm}
\end{table}
Building upon our observations, we identified three primary dimensions contributing to confusion induction: action variability, individual behavioral patterns, and explanation level modulation. Regarding individual behavioral patterns, confusion may not always be easily recognizable, yet its impact on interaction may be notable, often leading to increased negative emotions and decreased positive emotions \cite{detecting_confusion_int_2021,identify_int_confusion_LI2023,taxonomy_social_errorsin_HRI_jrHRI2021}. This insight suggests that confusion can elicit specific responses that are not always overt. In light of this, we want to develop a model that detects subtle indicators of confusion to predict whether a decrease or even maintaining the same explanation level leads to confusion induction. 



We define a predictor as below:
\begin{equation}
    y = f(A, X, D) = \begin{cases}
  0, & \text{NC (Not Confused)}\\
  1, & \text{C (Confused)}
\end{cases}
\label{eq:classification_Task}
\end{equation}
The predictor is a function of several input features (Table \ref{tab:features}), where A represents the robot’s failure action, X denotes the human’s last and current reaction to this action, and D is a decrease in the explanation level. The human reaction is quantified using the behavior metrics listed in Table \ref{tab:behavior_measures}. The predictor outputs whether the person will be in a state of confusion induction after receiving the robot’s explanation. 
\subsection{Prediction Task and Training Data}
To train the desired predictor for the Prediction task in Eq. 
\eqref{eq:classification_Task}, we utilize the recorded data as well as confusion annotations obtained via Algorithm 1. The dataset consists of the confusion states of 55 participants, each experiencing 8 failure cases. Although there were originally 11 failures for each participant, 3 from the first round were excluded due to the lack of previous reactions. Further data filtering led to a total of 439 samples with 20.5\% confused and 79.5\% not confused instances. By analyzing both historical and current interaction data, including the chosen action and changes in explanation levels, the trained model aims to predict confusion induction accurately.
\subsection{Training and Testing Procedure}
For our training and testing procedure, we employed a leave-one-participant-out cross-validation approach \cite{Multimodal_User_Feedback_AGNES2022,MHHRI_DATASET_humanhuman_oya2019}. This methodology involved 55 distinct cross-validation iterations, corresponding to the number of participants in our study. In each iteration, data from a single participant was isolated as the test set, while the model was trained on the collective data from the remaining 54 participants. This process was systematically repeated to ensure each participant's data was used exactly once as the test set. 

Our chosen approach offered two key advantages. First, it ensured that variations in behavior, confusion responses, or interaction styles between participants were not represented in both the training and test sets, thereby providing a more realistic evaluation scenario. Second, it illustrated whether their behaviors were similar to or different from those expressed by the other participants in the training data, offering insights into individual variability. Consequently, this approach  allowed for a robust assessment of the model's ability to generalize to  behaviors from entirely novel individuals.

\subsection{Model Selection and Evaluation}

To identify the optimal model for predicting confusion induction, accounting the variability in human behavior, we evaluated multiple machine learning algorithms (SVM, Neural Networks, Tree-based, etc). Model selection was guided by accuracy and F1-score metrics, with detailed results presented in \cite{git_paper_andreas}.
A particular emphasis was placed on minimizing false negatives, as these errors can impact the model’s ability to accurately detect confusion. To address this, we applied class weights during the training process and assigned higher weights to the Confused class to improve the model’s sensitivity to instances of confusion. Also, we fine-tuned the models by systematically adjusting various parameters with grid search to optimize performance. 


The best performing model, a Random Forest model (depth=10, split=5, leaf=10), yields results with an average accuracy of 89.54\% and an average precision of 0.878. This performance of tree-based algorithms, particularly Random Forests, is further supported by prior research \cite{Analysis_of_human_rxn_to_robot_convo_failures_DIMOS2021, Multimodal_User_Feedback_AGNES2022} in related domains, applying similar concepts successfully. 



\section{Adaptive Failure Explanation}
\subsection{Mechanism} 
We propose an approach where the robot starts with a high-level explanation and then adjusts it as needed, i.e. as user's mental model of the interaction and robot is complemented. This approach is based on the observation in \cite{exp_strategies_roman_khanna2023} that it is sufficient to provide a high level of detail initially and reduce this level in later stages. 
Consequently, we propose a mechanism for adjusting the explanation level (\(E\)) based on Algorithm 2. By utilizing this adjustment rule, the robot assesses whether to decrease, increase, or maintain the current level of explanation \(E_{\text{current}}\) based on real-time evaluations and determines the new explanation level denoted as \(E_{\text{new}}\). \(E\) is restricted to ensure it does not go below the minimum level \(E_{\text{min}}\) or exceed the maximum level \(E_{\text{max}}\). This mechanism specifies that \(E\) should be reduced if the predictor \(f\), with a decreased level of explanation (D=1), predicts that the human would not be confused. If not, the mechanism  specifies that \(E\) should be increased if (D=0) maintaining  same \(E_{\text{current}}\) would still confuse the participant. 
If neither of these conditions is valid, the mechanism opts to maintain the current level. This implies that the predictor indicates that the participant will be confused if the level decreases and that keeping the same level will prevent confusion. 


\begin{algorithm}[t]
\small
\caption{Decision Rule for the Explanation Level}
\begin{algorithmic}[1]
\State SET $D \gets 1$
\If{$f(A, X, D=1) = \text{``NC"}$}
    \State $E_{\text{new}} \gets \max(E_{\text{current}} - 1, E_{\text{min}})$   \quad \quad 
 (\textit{Decrease E})
\ElsIf{$f(A, X, D=1) = \text{``C"}$}
    \State SET $D \gets 0$
    \If{$f(A, X, D=0) = \text{``C"}$}
        \State $E_{\text{new}} \gets \min(E_{\text{current}} + 1, E_{\text{max}})$   \quad 
 (\textit{Increase E})
    \Else
        \State $E_{\text{new}} \gets E_{\text{current}}$
 (\textit{Same E})
    \EndIf
\EndIf
\end{algorithmic}
\end{algorithm}

\subsection{Mechanism Evaluation}
To assess the effectiveness of our mechanism, we applied it to our dataset and compared the mechanism’s decisions with participants’ actual confusion states. For each robotic failure, the mechanism suggested increasing, decreasing, or keeping same explanation level. 
The decisions were then categorized based on whether the mechanism’s recommendation was followed or not in our dataset, which resulted in 4 distinct cases: Suggested Increase - Not Followed, Suggested Same - Followed, Suggested Decrease - Followed, and Suggested Decrease - not Followed. We did not find the Suggested Same-Not Followed case in our analysis. Also, the dataset only included cases with either constant or decreased explanation levels, with no increases, as detailed in Table \ref{tab:round-results}.
Based on these distinct cases, we propose the following hypothesis to evaluate the adaptive failure explanation mechanism:
\begin{itemize}[leftmargin=*]
    \item H1: \textit{When the mechanism recommends increasing the explanation level and this recommendation is not followed, participants are more likely to be confused.}
    \item H2: \textit{When the mechanism recommends keeping the explanation level same and this recommendation is followed, participants are less likely to be confused.}.
    \item H3: \textit{When the mechanism recommends decreasing the explanation level, participants are less likely to be confused (irrespective of recommendation being followed or not).
    }
\end{itemize}
The null hypothesis for H1 and H2 is: There is no association between following the mechanism's recommendation and the participant's confusion; and for H3: There is no association between the mechanism's recommendation to decrease the explanation level and the participant's confusion. 

\begin{figure}[t]
      \centering
      \setlength\abovecaptionskip{-0.05\baselineskip}
      \includegraphics[width=1.0\linewidth,trim={0.3cm 0.5cm 0.01cm 0.2cm},clip]{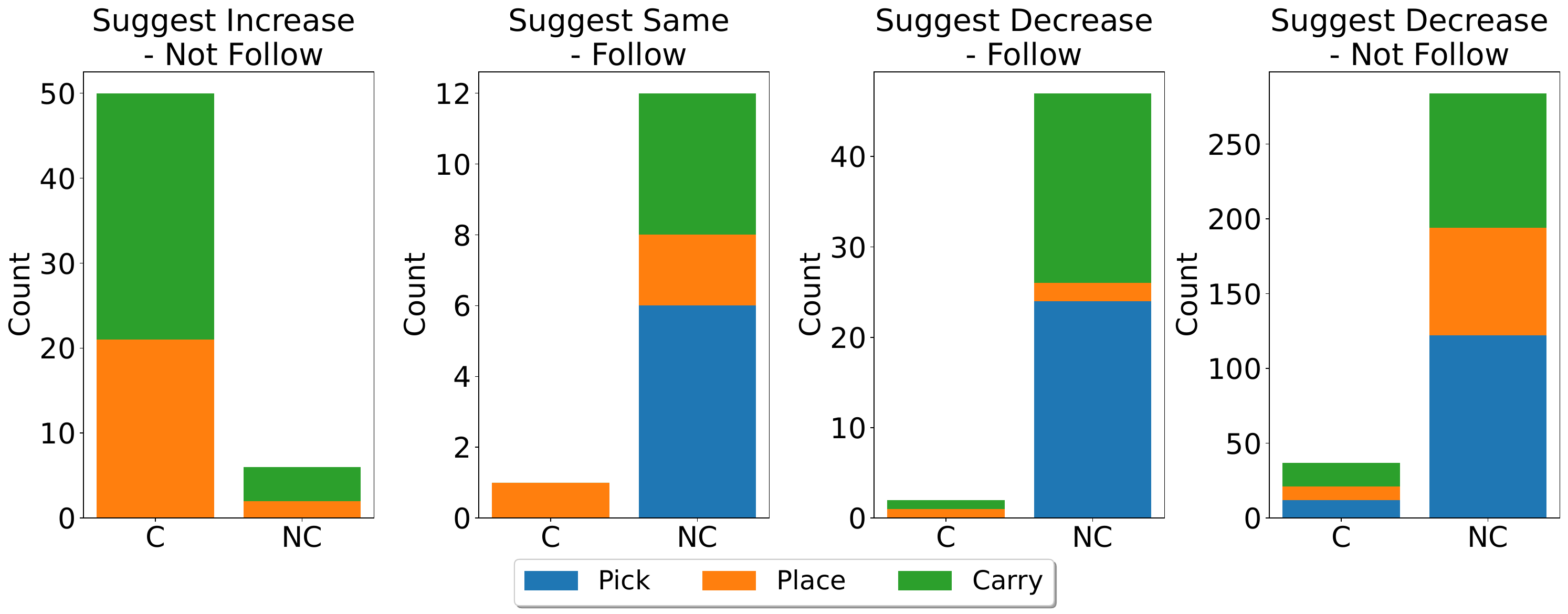}
       \caption{Mechanism Evaluation: Confusion induction across different tasks based on whether suggested explanation adjustments were followed or not.}
      \vspace{-5mm}
      \label{fig:mechanism_eval}
\end{figure}

We perform a chi-square test of independence to reject each null hypotheses. For each $\text{$H_i$}$, we compare the specific case against the overall distribution.
The results of this evaluation are summarized in Fig.\ref{fig:mechanism_eval} and the following:
\subsubsection{\textbf{Suggested Increase - Not Followed}}
There were a significant number of confusion instances (50 out of 56), indicating that following the recommendation could have mitigated confusion. From the chi-square test, we get significant results with $\chi^2$(1, N = 495) = 115.8466, p $<$  0.00001, rejecting the null hypothesis. Also, these results hold valid for place and carry failures as well. Hence H1 is found valid. 
\subsubsection{\textbf{Suggested Same - Followed}}
The majority did not result in confusion (12 out of 13). However, the results from the chi-square test are not significant, with $\chi^2$(1, N = 495) = 1.2883, p = 0.256. 
This suggests that while there are indications supporting H2, the current sample size is insufficient to draw a conclusive result.

\subsubsection{\textbf{Suggested Decrease}}
For cases when the mechanism suggested a decrease, the results from the chi-square test are found significant with $\chi^2$(1, N = 495) = 14.8623, p $<$ 0.005. Hence, H3 is found valid. 

\textbf{Suggested Decrease - Followed:} 
Most instances (47 out of 49) where the explanation level was reduced did not result in confusion. This outcome demonstrates the mechanism's effectiveness in identifying cases where a decrease in explanation level is appropriate. These results also hold valid across pick, carry, and place failures.

\textbf{Suggested Decrease - Not Followed:} 
The majority of failure instances (284) did not result in confusion, while a few instances (37) did. This suggests that, although the recommendation to lower the explanation level might have been appropriate, the existing explanation level was generally effective in preventing confusion for most instances. 

\subsection{Findings:}
Our proposed mechanism, which adjusts the level of explanation, demonstrates that strategically managing explanation levels could reduce confusion and improve interaction quality in HRC scenarios with recurring failures. Specifically, when the mechanism recommended an increase in explanation level but was not followed (H1), there was a significant rise in confusion, suggesting that following the recommendation could have mitigated this effect. Conversely, when the mechanism recommended decreasing the level of explanation (H3), most instances did not result in confusion, showing that the mechanism effectively identifies when less detail is appropriate. Lastly, while we found that maintaining the same level of explanation generally led to reduced confusion (H2), further data is needed to verify its impact. 


\section{Discussion, Limitations and Future work}


This work aimed to emphasize the critical need for computational interpretations of human behavior based on social cues and multimodal approach. By adapting communication based on real-time evaluations, robots can align their behavior with the user’s cognitive and mental state, resulting in the user’s sense of being understood. Identifying cognitive and affective metrics, such as confusion, can also provide deeper insights into interaction patterns in HRC. Ultimately, advancing our understanding of how adaptive communication, such as our proposed mechanism, enhances trust and rapport will contribute to the development of more natural HRC. 

Additionally, our study suggests that existing HRC strategies could benefit from incorporating confusion as a key design variable. Monitoring confusion states/levels could reveal important shifts in trust, engagement, and emotional responses, enabling the development of adaptive systems that can proactively address confusion-related challenges. By demonstrating how confusion can be integrated into robot design, our approach offer a novel perspective on HRC.
While our study offers valuable insights, these limitations should be acknowledged:

\textbf{Methodological Challenges in Confusion Assessment:}
Research on confusion in HRI is still evolving, presenting significant methodological challenges. Due to the absence of an established theoretical framework for identifying confusion states, we were unable to rely on standardized assessment methods. The subjective nature of confusion, coupled with potential limitations in self-reporting \cite{detecting_confusion_int_2021} and manual annotation, led us to implement an automated assessment approach. However, our approach could benefit from further validation and theoretical refinement. We encourage future research to develop reliable confusion assessment methodologies, and explore objective indicators of confusion in HRI.


\textbf{Limitations of Context-Specific Scenarios:}  
A key limitation of our study is its reliance on lab-collected data and specific use cases, which may restrict the generalizability of our approach. Focusing on specific failure scenarios may limit the applicability of our findings to broader HRC contexts. To overcome these limitations, we advocate for the integration of multimodal systems and foundational models to develop more adaptable and broadly applicable frameworks.




\textbf{Engagement and Information Length:} 
The length of failure explanation information varied across levels, which may have influenced participant interactions and engagement. Future studies could control for explanation length by introducing alternative delivery formats, such as allowing users to request additional details as needed. 


\textbf{Limited Scope of Confusion Stimuli:}  
Our study primarily examined confusion induced by insufficient information, however, this represents only one of the confusion-inducing factors (complex or contradictory information). We encourage future research to explore a wider range of confusion stimuli, allowing for a more comprehensive understanding of how different types of confusion shape HRC.


\section{Conclusion}
In this paper, we presented an automated approach for confusion induction assessment, a data-driven approach for predicting confusion induction and an adaptive mechanism for failure explanation based on observed human behavior. We detailed our data collection and analysis procedures, and investigated the feasibility of building a model capable of predicting confusion. The performance of the proposed predictor were evaluated using accuracy and F1-score metrics, and an adaptive mechanism for adapting the failure explanation level based on observed human behavior. Finally, the evaluation shows that the mechanism is capable of adjusting explanation levels effectively to reduce confusion in many cases, while also lowering the level when appropriate. 

\bibliography{IEEE}
\bibliographystyle{IEEEtran}

\end{document}